\documentclass[nonacm, acmtog]{acmart}

\usepackage{amsmath,amsfonts}
\usepackage{algorithmic}
\usepackage{algorithm}
\usepackage{array}
\usepackage[caption=false,font=normalsize,labelfont=sf,textfont=sf]{subfig}
\usepackage{textcomp}
\usepackage{stfloats}
\usepackage{url}
\usepackage{verbatim}
\usepackage{graphicx}
\usepackage{layouts}

\usepackage{bbm}
\usepackage{amstext}
\usepackage{amsmath}
\usepackage{xcolor}
\usepackage[format=plain,labelformat=simple,labelsep=period,font=small,compatibility=false]{caption}
\usepackage{cleveref}
\usepackage{xspace}

\newcommand{\clean}{clean}

\newcommand{\methodname}{SpotLessSplats\xspace}
\newcommand{\methodshortname}{SLS\xspace}
\newcommand{\methodsdmlp}{\methodshortname-mlp\xspace}
\newcommand{\methodsdcluster}{{\methodshortname}-agg\xspace}
\newcommand{\methodprune}{UBP\xspace}
\newcommand{\methodfilter}{RobustFilter\xspace}

\newcommand{\tdgsshortname}{3DGS\xspace}

\newcommand{\mean}{\mu}
\newcommand{\covariance}{\boldsymbol{\Sigma}}
\newcommand{\opacity}{\alpha}
\newcommand{\sphcoef}{\mathbf{c}}
\newcommand{\viewxform}{\mathbf{W}}
\newcommand{\projjacobian}{\mathbf{J}}
\newcommand{\rot}{\mathbf{R}}
\newcommand{\scale}{\mathbf{S}}
\DeclareMathOperator*{\argmin}{arg\,min}
\newcommand{\image}{\mathbf{I}}
\newcommand{\nImages}{N}
\newcommand{\model}{\mathcal{G}}
\newcommand{\element}{g}
\newcommand{\latent}{\mathbf{z}}
\newcommand{\mlp}{\mathcal{H}}
\newcommand{\mlpparam}{\theta}

\newcommand{\features}{\mathbf{F}}

\newcommand{\mask}{\mathbf{M}}
\newcommand{\upperm}{\mathbf{U}}
\newcommand{\lowerm}{\mathbf{L}}

\newcommand{\step}{\scalebox{0.8}{(t)}}
\newcommand{\steps}{T}
\newcommand{\residual}{\mathbf{R}}
\newcommand{\real}{\mathbb{R}}
\newcommand{\indicator}{\mathbbm{1}}
\newcommand{\filter}{\mathbf{B}}

\newcommand{\loss}{\mathcal{L}}
\newcommand{\sched}{\alpha}
\newcommand{\maxx}{\text{max}}
\newcommand{\x}{x}
\newcommand{\util}{u}
\newcommand{\pthresh}{\kappa}
\newcommand{\aglolatent}{z}
\newcommand{\aglonet}{\mathcal{Q}}
\newcommand{\cluster}{\mathbf{C}}
\newcommand{\numclusters}{C}

\definecolor{cvprblue}{rgb}{0.21,0.49,0.74}
\definecolor{gold}{rgb}{0.7, 0.5, 0}

\newcommand{\at}[1]{#1}

\newcommand{\lpips}{\scalebox{0.8}{LPIPS$\downarrow$}}
\newcommand{\ssim}{\scalebox{0.8}{SSIM$\uparrow$}}
\newcommand{\psnr}{\scalebox{0.8}{PSNR$\uparrow$}}

\renewcommand{\paragraph}[1]{\vspace{.2em}\noindent\textbf{#1}.}

\usepackage{soul}
\setuldepth{foobar}

\usepackage{multirow}

\AtBeginDocument{%
  }

\setcopyright{none}
\copyrightyear{none}
\acmYear{none}
\acmDOI{none}

\begin{document}

\title{\methodname: Ignoring Distractors in 3D Gaussian Splatting
}
\author{Sara Sabour}
\authornote{Both authors contributed equally to this research.}
\email{sasabour@google.com}
\orcid{1234-5678-9012}
\author{Lily Goli}
\authornotemark[1]
\email{lily.goli@mail.utoronto.ca}
\affiliation{%
  \institution{Google Deepmind, University of Toronto}
  \city{Toronto}
  \country{Canada}
}

\author{George Kopanas}
\affiliation{%
  \institution{Google AR}
  \city{London}
  \country{United Kingdom}}

\author{Mark Mathews}
\affiliation{%
  \institution{Google Deepmind}
  \city{Mountain View}
  \country{United States}}

\author{Dmitry Lagun}
\affiliation{%
  \institution{Google Deepmind}
  \city{Mountain View}
  \country{United States}}

\author{Leonidas Guibas}
\affiliation{%
  \institution{Google Deepmind, Stanford University}
  \city{Mountain View}
  \country{United States}}

\author{Alec Jacobson}
\affiliation{%
  \institution{University of Toronto}
  \city{Toronto}
  \country{Canada}}

\author{David Fleet}
\affiliation{%
  \institution{Google Deepmind, University of Toronto}
  \city{Toronto}
  \country{Canada}}
\email{larst@affiliation.org}

\author{Andrea Tagliasacchi}
\affiliation{%
  \institution{Google Deepmind, University of Toronto, Simon Fraser University}
  \city{Vancouver}
  \country{Canada}}
\renewcommand{\shortauthors}{Sabour \& Goli et al.}

\begin{abstract}

3D Gaussian Splatting (3DGS) is a promising technique for 3D reconstruction, offering efficient training and rendering speeds, making it suitable for real-time applications. However, current methods require highly controlled environments—no moving people or wind-blown elements, and consistent lighting—to meet the inter-view consistency assumption of 3DGS. This makes reconstruction of real-world captures {problematic}. We present SpotLessSplats, an approach that leverages pre-trained and general-purpose features coupled with robust optimization to effectively ignore transient distractors.
Our method achieves state-of-the-art reconstruction quality both visually and quantitatively, on casual captures. Additional results available at: \url{https://spotlesssplats.github.io}
\end{abstract}

\begin{teaserfigure}
\includegraphics[width=\textwidth]{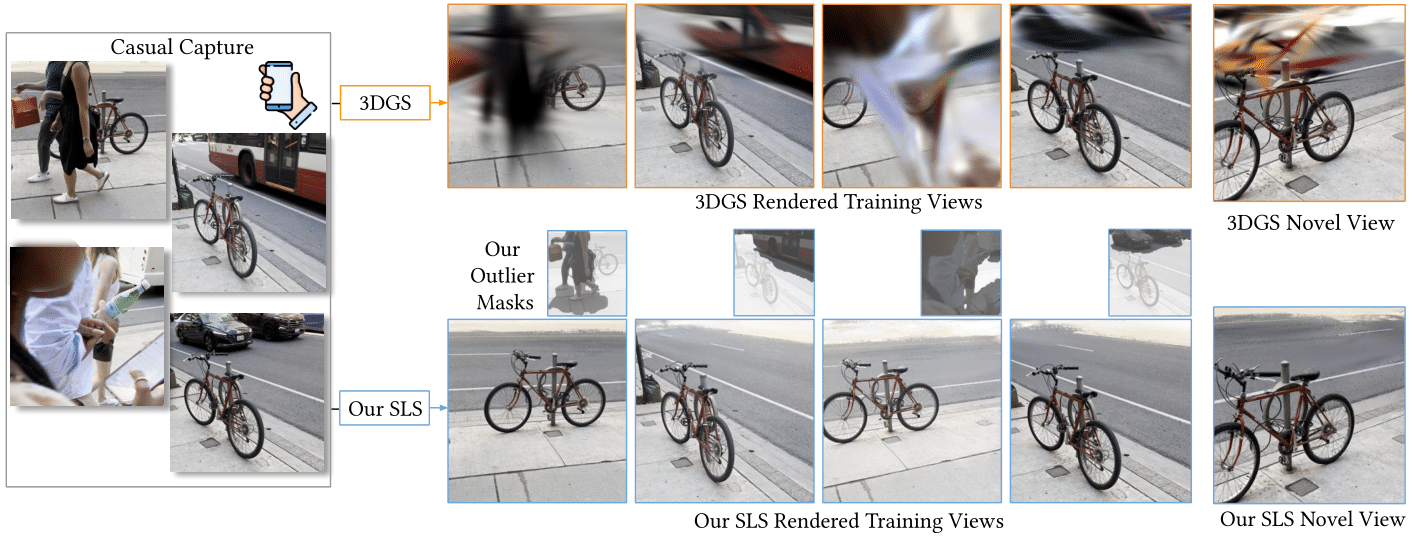}
\caption{{\methodname} cleanly reconstructs a scene with many transient occluders (\textbf{middle}), while avoiding artifacts~(\textbf{bottom}).
It correctly identifies and masks out all transients~(\textbf{top}), even in captures with a large number of them~(\textbf{left}).}
\label{fig:teaser}
\end{teaserfigure}

\maketitle

\section{Introduction}
\label{sec:introduction}

\begin{teaserfigure}
\includegraphics[width=\textwidth]{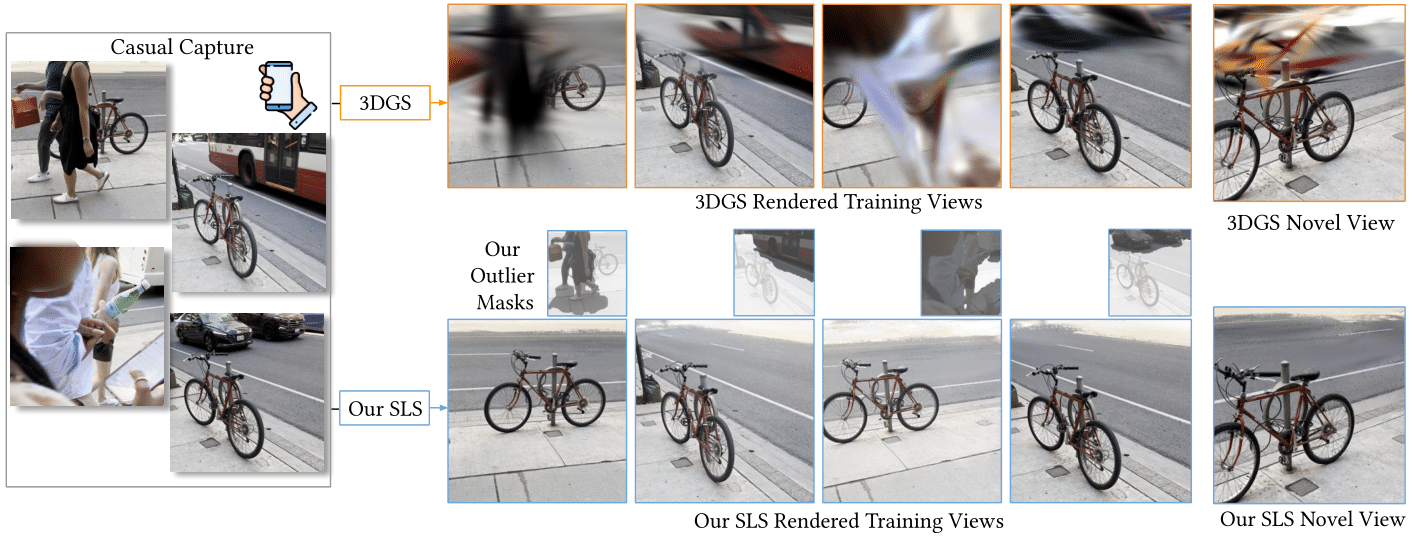}
\caption{{\methodname} cleanly reconstructs a scene with many transient occluders (\textbf{middle}), while avoiding artifacts~(\textbf{bottom}).
It correctly identifies and masks out all transients~(\textbf{top}), even in captures with a large number of them~(\textbf{left}).}
\label{fig:teaser}
\end{teaserfigure}

The reconstruction of 3D scenes from 2D images with neural radiance fields (NeRF)~\cite{Mildenhall20eccv_nerf} and, more recently, with 3D Gaussian Splatting (3DGS)~\cite{Kerbl2023tdgs}, has been the subject of intense focus in vision research.
Most current methods assume that images are simultaneously captured, perfectly posed, and noise-free.
While these assumptions simplify 3D reconstruction, they rarely hold in real-world, where moving objects (e.g., people or pets), lighting variations, and other spurious photometric inconsistencies degrade performance, limiting widespread application.

In NeRF training, robustness to outliers has been incorporated by 
down-weighting or discarding inconsistent observations based on the magnitude of color residuals \cite{Wu2022d2nerf, Sabour2023robustnerf, MartinBrualla21cvpr_nerfw, chen2024nerfhugs}.
Similar methods adapted to 3DGS \cite{Dahmani2024swag, Kulhanek2024wildgaussians, Wang2024wegs} address global appearance changes and single-frame transients seen in datasets like Phototourism~\cite{phototourism2006}.
Such captures include appearance changes occurring over weeks and different times of day, which are not common in most casual captures.
For 3DGS in particular, the \textit{adaptive densification} process itself introduces variance in color residuals, compromising detection of transients when directly applying existing ideas from robust NeRF frameworks.

In this paper we introduce \methodname~(\methodshortname), a framework for robust 3D scene reconstruction with 3DGS, {via unsupervised detection of outliers in training images}.
Rather than detecting outliers in RGB space, we instead utilize a richer, \textit{learned feature space} from text-to-image models.
The meaningful semantic structure of this feature embedding allows one to more easily detect the spatial support of structured outliers associated, for example, with a single object.
Rather than employing manually-specified robust kernels for outlier identification~\cite{Sabour2023robustnerf}, we instead exploit adaptive methods in this feature space to detect outliers.
To this end we consider two approaches within this framework.
The first uses  non-parametric clustering of local feature embeddings as a simple way to find image regions of structured outliers.
\at{The second uses an MLP, trained in an unsupervised fashion to predict the portion of the feature space that is likely to be associated with distractors.}
We further introduce a (complementary and general purpose) sparsification strategy, compatible with our robust optimization, that delivers similar reconstruction quality with two to four times fewer splats, even on distractor-free datasets, yielding significant savings in compute and memory.
Through experiments on challenging benchmarks of casually captured scenes~\cite{Sabour2023robustnerf, Ren2024nerfonthego}, \methodshortname is shown to consistently outperform competing methods in reconstruction accuracy.

\vspace{1em}

\noindent
Our key contributions include:
\begin{itemize}
    \item {An adaptive, robust loss, leveraging text-to-image diffusion features, that reliably identifies transient distractors in causal captures, eliminating issues of overfitting to photometric errors.}
    \item A novel sparsification method compatible with our robust loss that significantly reduces the number of Gaussians, saving compute and memory without loss of fidelity.
    \item Comprehensive evaluation of \methodshortname on standard benchmarks, demonstrating SOTA robust reconstruction, outperforming existing methods by a substantial margin.
\end{itemize}
\section{Related work}
\label{sec:related}

Neural Radiance Fields (NeRF) \cite{Mildenhall20eccv_nerf}, have gained widespread attention due to the high quality reconstruction and novel view synthesis of 3D scenes.
NeRF represents the scene as a view dependent emissive volume.
The volume is rendered using the absorption-emission part of the volume rendering equation~\cite{Kajiya1984rtv}.
Multiple enhancements have followed.
Fast training and inference~\cite{Sun2022dvgo, Muller2022ingp, Yu21iccv_PlenOctrees, mobilenerf}, training with limited or single view(s)~\cite{Yu21cvpr_pixelNeRF, Jain21iccv_DietNeRF, Rebain22lolnerf} and simultaneous pose inference~\cite{Lin21iccv_BARF, Wang21arxiv_NeRFminusminus, Levy2024melon} have brought radiance fields closer to practical applications.
More recently, 3D Gaussian Splatting (3DGS)~\cite{Kerbl2023tdgs} was proposed as a primitive-based alternative to NeRFs with significantly faster rendering speed, while maintaining high quality.
3D Gaussians can be efficiently rasterized using alpha blending~\cite{Zwicker2001splatting}. This simplified representation takes advantage of modern GPU hardware to facilitate real-time rendering.
\at{The efficiency and simplicity of 3DGS have prompted a shift in focus within the field, with many NeRF enhancements being quickly ported to 3DGS~\cite{Yu2024MipSplatting, charatan2024PixelSplat}.}

\begin{figure}[t]
\centering
\includegraphics[width=\linewidth]{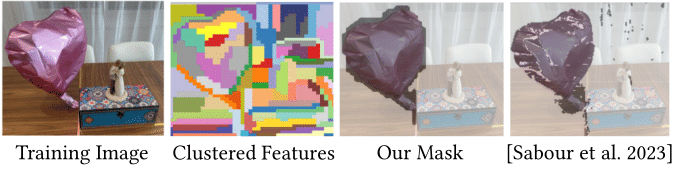}
\captionof{figure}{
Our outlier classification using clustered semantic features covers the distractor balloon fully, but an adapted robust mask from~\cite{Sabour2023robustnerf} misclassifies pixels with similar color to background, as inliers.
}
\label{fig:clustering}
\end{figure}
\paragraph{Robustness in NeRF}
The original NeRF paper made strong assumptions regarding the capture setup: the scene needs to be perfectly static, and the illumination should stay unchanged throughout the capture.
More recently, NeRF has been extended to train on unstructured~``in-the-wild'' captured images that violate these constraints.
Two influential works, NeRF-W~\cite{nerfw} and RobustNeRF~\cite{Sabour2023robustnerf} addressed the problem of transient distractors, both using photometric error as guidance.
NeRF-W~\cite{nerfw} models a 3D uncertainty field rendered to 2D outlier masks that down-weight the loss at pixels with high-error, and a regularizer that prevents degenerate solutions.
NeRF-W~\cite{nerfw} also models global appearance via learned embeddings, which are useful for images captured over widely varying lighting and atmospheric conditions. 
Urban Radiance Fields (URF)~\cite{Rematas2022urf} and Block-NeRF~\cite{Tancik2022blocknerf} similarly apply learned appearance embeddings to large-scale reconstruction.
HA-NeRF~\cite{Chen2022hanerf} and Cross-Ray~\cite{Yang2023crossray} model 2D outlier masks instead of 3D fields, leveraging CNNs or transformers for cross-ray correlations.

RobustNeRF~\cite{Sabour2023robustnerf}, approached the problem from a robust estimator perspective, with binary weights determined by thresholded rendering error, \at{and a blur kernel to reflect the assumption that pixels belonging to distractors are spatially correlated.}
However, both RobustNeRF and NeRF-W variants~\cite{Chen2022hanerf, Yang2023crossray} rely solely on RGB residual errors and because of this they often misclassify transients with colors similar to their background; see RobustMask in~\Cref{fig:clustering}.
To avoid this, previous methods require  careful tuning of hyper-parameters, \at{i.e., the blur kernel size and thresholds in RobustNeRF and the regularizer weight in NeRF-W}.
On the contrary, our method uses the rich representation of text-to-image models for semantic outlier modeling. This avoids direct RGB error supervision, as it relies on feature-space similarities for clustering.

\begin{figure}[t]
\centering
\setlength\tabcolsep{1.8pt}
\includegraphics[width=\linewidth]{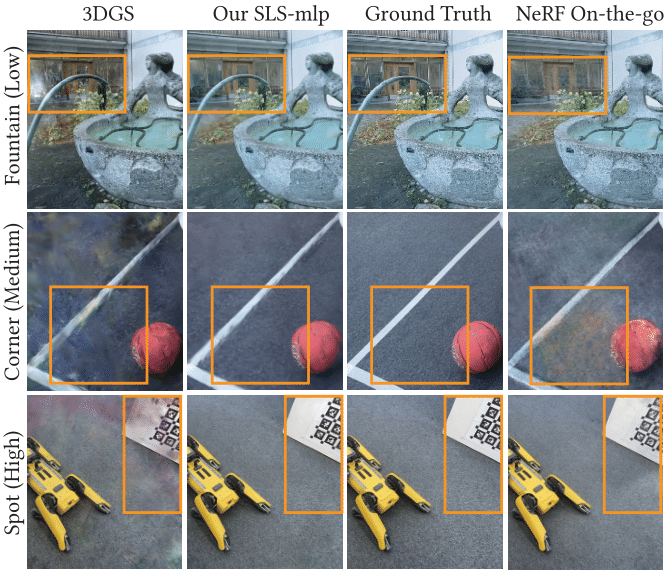}\\

\resizebox{\linewidth}{!}{

\begin{tabular}{l|ccc|ccc|ccc}
 & \multicolumn{3}{c|}{Fountain} & \multicolumn{3}{c|}{Corner} & \multicolumn{3}{c}{Spot}\\
 & \psnr & \ssim & \lpips & \psnr & \ssim & \lpips & \psnr & \ssim & \lpips  \\ \hline
MipNerf360 & 13.91 & 0.29 & 0.55 & 20.41 & 0.66 & 0.34  & 17.82 & 0.30 & 0.46 \\
RobustNerf & 17.20 & 0.41 & 0.54 & 20.21 & 0.70 & 0.35 & 16.40 & 0.38 & 0.69 \\
NeRF On-the-go & 20.11 & 0.61 & 0.31 & 24.22 & 0.80 & 0.19 & 23.33 & 0.79 & 0.19 \\ \hline
\tdgsshortname & 21.70 & 0.79 & 0.16 & 24.05 & 0.86 & 0.13 & 20.72 & 0.76 & 0.31 \\
Our \methodsdmlp & \textbf{22.81} & \textbf{0.80} & \textbf{0.15} & \textbf{26.43} & \textbf{0.90} & \textbf{0.10} & \textbf{25.76} & \textbf{0.90} & \textbf{0.12}
\end{tabular}
} %
\captionof{figure}{
Our method accurately reconstructs scenes with different levels of transient occlusion, avoiding leakage of transients or under-reconstruction evident by the quantitative and qualitative results on NeRF On-the-go~\cite{Ren2024nerfonthego} dataset.
}
\label{fig:nerfonthego1}
\end{figure}

NeRF On-the-go~\cite{Ren2024nerfonthego} released a dataset of casually captured videos with transient occluders. 
Similar to our method, it uses semantic semantic features from DINOv2~\cite{Oquab2023dinov2} to predict outlier masks via a small MLP. 
However, it also relies on direct supervision from the structural rendering error, leading to potential over- or under-masking of outliers. 
This is illustrated in~\Cref{fig:nerfonthego1}, where over-masking has removed the hose (`Fountain') and has smoothed the carpet (`Spot'), while under-masking caused distractor leaks and foggy artifacts (`Corner' and `Spot').
NeRF-HuGS~\cite{chen2024nerfhugs} combines heuristics from COLMAP's robust sparse point cloud~\cite{schoenberger2016sfm}, and off-the-shelf semantic segmentation to remove distractors.
Both heuristics are shown to fail under heavy transient occlusions in~\cite{Ren2024nerfonthego}.

\paragraph{Precomputed features}
The use of precomputed vision features, such as DINO~\cite{Caron2021dino, Oquab2023dinov2} have demonstrated the ability to generalize to multiple vision tasks.
Denoising Diffusion Probabalistic Models~\cite{song2019generative, ho2020denoising, Rombach2022stablediffusion}, known for their photorealistic image generation capabilities from text prompts~\cite{saharia2022photorealistic,ramesh2022hierarchical,rombach2021high}, have been shown to have internal features similarly powerful in generalizing over many tasks e.g. segmentation and keypoint correspondence~\cite{Amir2022deepvit, Tang2023emergent, Hedlin2024stablekeypoints, Zhang2023tale, Luo2023diffhyperfeatures}.

\paragraph{Robustness in 3DGS (concurrent works)}
Multiple concurrent works address 3DGS training on wild-captured data. 
SWAG~\cite{Dahmani2024swag} and GS-W~\cite{Zhang2024gsw} model appearance variation using learned global and local per-primitive appearance embeddings.
Similarly, WE-GS~\cite{Wang2024wegs} uses an image encoder to learn adaptations to the color parameters of each splat, per-image. Wild-GS~\cite{xu2024wildgs} learns a spatial triplane field for appearance embeddings.
All such methods~\cite{Zhang2024gsw, Wang2024wegs, xu2024wildgs} adopt an approach to outlier mask prediction like NeRF-W~\cite{nerfw}, with 2D outlier masks predicted to downweight high-error rendered pixels. SWAG \cite{Dahmani2024swag} learns a per-image opacity for each Gaussian, and denotes primitives with high opacity variance as transients.
{Notable are SWAG~\cite{Dahmani2024swag} and GS-W\cite{Zhang2024gsw} that show no or little improvement over the local/global appearance modeling, when additional learned transient masks are applied to Phototourism scenes~\cite{phototourism2006}.}
SLS focuses on casual captures with longer duration transients and minimal appearance changes, common in video captures like those in the ``NeRF on-the-go'' dataset~\cite{Ren2024nerfonthego}.

\section{Background}
\label{sec:background}
We build our technique on top of 3D Gaussian Splatting~\cite{Kerbl2023tdgs}, or 3DGS for brevity, which represents a 3D scene as a collection of 3D anisotropic Gaussians~$\model{=} \{ \element_i \} $, henceforth referred to as splats.
Given a set of posed images $\{ \image_n \}_{n = 1}^{\nImages}$,  $\image_i \in \real^{H \times W}$ of a casually captured scene, we aim to learn a 3DGS reconstruction $\model$ of the scene.
Each splat $\element_i$, is defined by a mean $\mean_i$, a positive semi-definite covariance matrix $\covariance_i$, an opacity $\opacity_i$, and view dependent color parameterized by spherical harmonics coefficients~$\sphcoef_i$~\cite{Ramamoorthi2001sph}.

The 3D scene representation is rendered to screen space by rasterization.
The splat positions/means are rasterized to screen coordinates via classical projective geometry, while special care needs to be taken to rasterize the covariance matrix of each splat.
In particular, if we denote with $\viewxform$ the perspective transformation matrix, the projection of the 3D covariance to 2D screen space can be approximated following~\cite{Zwicker2001splatting} as $\tilde\covariance = \projjacobian \viewxform \covariance \viewxform^{T} \projjacobian^{T}$,
where $\projjacobian$ is the Jacobian of the projection matrix, which provides a linear approximation to the non-linear projection process.
To ensure $\covariance$ represents covariance throughout optimization~(i.e., positive semi-definite), the covariance matrix is parameterized as $\covariance = \rot\scale\scale^{T}\rot^{T}$,
where scale $\scale{=}\text{diag}(\mathbf{s})$ with $\mathbf{s}{\in}\real^3$, and rotation $\rot$ is computed from a unit Quaternion $q$.
Once splat positions and covariances in screen-spaces are computed, the image formation process executes volume rendering as alpha-blending, which in turn requires splat sorting along the view direction.
Unlike NeRF, which renders \textit{one pixel at a time}, 3DSG renders the \textit{entire image} in a single forward pass.

\subsection{Robust optimization of 3DGS}
\label{sec:robust}
Unlike typical capture data for 3DGS~\cite{Kerbl2023tdgs}, we do not assume the set of posed images~$\{ \image_n \}_{n = 1}^{\nImages}$ to be curated, but rather \textit{casually captured}.
That is, we \textit{do not} require images to be depictions of a perfectly 3D consistent and static world.
Following prior work, we (interchangeably) denote the portion of images that break these assumptions as \textit{distractors}~\cite{Sabour2023robustnerf} or \textit{transient effects}~\cite{MartinBrualla21cvpr_nerfw}.
And unlike prior works~\cite{kerbl2024hierarchical,MartinBrualla21cvpr_nerfw,Tancik2022blocknerf}, we do not make assumptions about the transient object class, appearance and/or shape.

We address this problem by taking inspiration from the pioneering work of~\cite{Sabour2023robustnerf} in RobustNeRF, which removes distractors by identifying the portion of input images that should be masked out in the optimization process.
The problem reduces to predicting (without supervision) inlier/outlier masks $\{ \mask_n \}_{n = 1}^{\nImages}$ for each training image, and optimizing the model via a \textit{masked} L1 loss:
\begin{equation}
    \argmin_{\model} \sum_{n=1}^{\nImages}  \mask_{n}^{\step} \odot
    \| \image_{n} - \hat{\image}_{n}^{\step} \|_1.
    \label{eq:nvs}
\end{equation}
where $\hat{\image}_{n}^{\step}$ is a rendering of $\model$ at training iteration~$(t)$.
As in RobustNeRF~\cite{Sabour2023robustnerf}, transient effects can be detected by observing photometric inconsistencies during training; that is, image regions that are associated with a large loss value.
By denoting with $\residual^{\step}_n {=} \|\image_{n} - \hat{\image}_{n}^{\step}\|_1$ the image of residuals~(with a slight abuse of notation, as the 1-norm is executed \textit{pixel-wise}, along the color channel), the mask is computed as:
\begin{align}
    \mask_{n}^{\step} {=} \indicator\left\{\left (\indicator\{\residual_{n}^{\step} {>} \rho\} \circledast \filter \right) {>} 0.5 \right\}
    \!,\:
    P(\residual_{n}^{\step}{>}\rho) {=} \tau
    \label{eq:robustnerf}
\end{align}
where $\indicator$ is an indicator function returning $1$ if the predicate is true and $0$ otherwise, $\rho$ is a generalized median with $\tau$ being a hyper-parameter controlling the cut-off percentile\footnote{If $\tau{=}.5$ then $\rho{=}\text{median}(\residual_{n}^{\step})$}, and $\filter$ is a~(normalized) $3\times3$ box filter that performs a morphological dilation via convolution~($\circledast$).
Intuitively, RobustNeRF~\cite{Sabour2023robustnerf}, summarized by \cref{eq:robustnerf} above, extends a \textit{trimmed robust estimator}~\cite{trimmedicp} by assuming that inliers/outliers are \textit{spatially correlated}.
We found that directly applying ideas from~\cite{Sabour2023robustnerf} to 3DGS, even when not limited by cases of misleading color residual like those depicted in~\Cref{fig:clustering}, do not remove outliers effectively.
Rather, several adaptations are necessary in order to accommodate differences in the representation and training process of 3DGS; see~\Cref{sec:adaptations}.

\section{Method}
\label{sec:method}
The outlier mask in~\cref{eq:robustnerf} is built solely based on photometric errors in the novel view synthesis process.
Conversely, we propose to identify distractors based on their semantics, \textit{recognizing} their re-occurrence during the training process.
We consider semantics as \textit{feature maps} computed from a self-supervised 2D foundation model (e.g.~\cite{Tang2023emergent}).
The process of removing distractors from training images then becomes one of \ul{identifying the sub-space of features that are likely to cause large photometric errors}.
As an example, consider a dog walking around in an otherwise perfectly static scene.
We would like to design a system that {either spatially in each image (\cref{sec:agglomerative}) or more broadly, spatio-temporally in the dataset (\cref{sec:learnt})}, recognizes ``dog'' pixels as the likely cause of reconstruction problems, and automatically removes them from the optimization. 
Our method is designed to reduce  reliance on local color residuals for outlier detection and over-fitting to color errors, and instead emphasizing reliance on semantic feature similarities between pixels. We thus refer to our methods as ``clustering.''
In~\Cref{sec:semantic} we detail how to achieve this objective. 
In~\Cref{sec:adaptations} we then detail several key adjustments to adapt the ideas from RobustNeRF~\cite{Sabour2023robustnerf} to a 3DGS training regime; see~\Cref{sec:agglomerative,sec:learnt}.

\subsection{Recognizing distractors}
\label{sec:semantic}
Given the input images $\{ \image_n \}_{n = 1}^{\nImages}$, we pre-compute feature maps for each image using Stable Diffusion~\cite{Rombach2022stablediffusion} as proposed by~\cite{Tang2023emergent}, resulting in feature maps~$\{\features_n\}_{n=1}^{\nImages}$.
This pre-processing step is executed \textit{once} before our training process starts.
We then employ these feature maps to compute the inlier/outlier masks~$\mask^{\step}$; we drop the image index $n$ to simplify notation, as the training process involves one image per batch.
We now detail two different ways to detect outliers.

\subsubsection{Spatial clustering}
\label{sec:agglomerative}
In the pre-processing stage, we additionally perform  unsupervised clustering of image regions.
Similar to super-pixel techniques {\cite{li2015superpixel,ibrahim2020image}}, we over-segment the image into a fixed cardinality collection of $\numclusters$ spatially connected components; see `Clustered Features'~\cref{fig:clustering}.
In more detail, we execute agglomerative clustering~\cite{mullner2011agglomerative} on the feature map~$\features$, where each pixel is connected to its $8$ surrounding pixels.
We denote the clustering assignment of pixel $p$ into cluster $c$ as~$\cluster[c,p]{\in}\{0,1\}$, and clustering is initialized with every pixel in its own cluster.
Clusters are agglomerated greedily, collapsing those that cause the least amount of inter-cluster feature variance differential before/post collapse.
Clustering terminates when $\numclusters{=}100$ clusters remain.

We can then calculate the probability of cluster $c$ being an inlier from the percentage of its inlier pixels in \cref{eq:robustnerf}:
\begin{equation}
P(c \in \mask^{\step}) = \Bigl( \sum_p \cluster[c,p] \cdot \mask^{\step}[p] \Bigr) ~/~ {\sum_p \cluster[c,p]},
\label{eq:agglomerativeprob}
\end{equation}
and then propagate the cluster labels back to pixels as:
\begin{equation}
\mask^{\step}_\text{agg}(p) = \sum_c \indicator\{P(c \in \mask^{\step}) > 0.5\} \cdot \cluster[c,p]
\label{eq:agglomerative}
\end{equation}
We then use $\mask^{\step}_\text{agg}$, rather than $\mask^{\step}$, as inlier/outlier mask to train our 3DGS model in~\cref{eq:nvs}.
We designate this model configuration as `\methodsdcluster'.

\subsubsection{Spatio-temporal clustering}
\label{sec:learnt}

A second approach is to train a \textit{classifier} that determines 
whether or not pixels should be included in the optimization~\cref{eq:nvs}, based on their associated features.
To this end we use an MLP with  parameters~$\mlpparam$ that predicts pixel-wise inlier probabilities from pixel features:
\begin{equation}
\mask^{\step}_\text{mlp} = \mlp(\features; \mlpparam^{\step}).
\label{eq:learnt}
\end{equation}
As the $\mlpparam^{\step}$ notation implies, the classifier parameters are updated \textit{concurrently} with 3DGS optimization.
$\mlp$ is implemented with $1 {\times} 1$ convolutions, and hence acts in an i.i.d.\  fashion across pixels. 
We interleave the optimization of the MLP and the 3DGS model, such that the parameters of one are fixed while the other's are optimized, in a manner similar to alternating optimization.

\begin{figure}[t]
\centering
\setlength\tabcolsep{1.8pt}
\includegraphics[width=\linewidth]{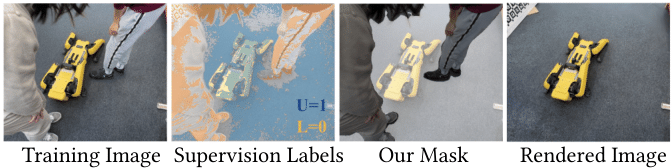}
\vspace*{-0.55cm}
\captionof{figure}{
Lower and upper error residual labels provide a weak supervision for training an MLP classifier for detecting outlier distractors.
}
\vspace*{-0.15cm}
\label{fig:mlp}
\end{figure}

The MLP classifier loss is given by
\begin{equation}
\loss(\mlpparam^{\step}) =  \loss_{sup}(\mlpparam^{\step}) + \lambda \loss_{reg}(\mlpparam^{\step}),
\label{eq:mlploss}
\end{equation}
with $\lambda{=}0.5$, and where $\loss_{sup}$ supervises the classifier:
\begin{align}
   \loss_{sup}(\mlpparam^{\step}) &= \maxx (\upperm^{\step} - \mlp(\features; \mlpparam^{\step}), 0)  \\ 
   &+ \maxx ( \mlp(\features; \mlpparam^{\step}) - \lowerm^{\step}, 0) \nonumber
\end{align}
and $\upperm$ and $\lowerm$ are self-supervision labels computed from the mask of the current residuals:
\begin{align}
\upperm^{\step} &= \mask^{\step} ~\text{from \cref{eq:robustnerf} with}~ \tau=.5
\\
\lowerm^{\step} &= \mask^{\step} ~\text{from \cref{eq:robustnerf} with}~ \tau=.9 
\end{align}
In other words, we directly supervise the classifier only on pixels for which we can confidently determine the inlier status based on reconstruction residuals, and otherwise we heavily rely on semantic similarity in the feature space; see \Cref{fig:mlp}.
To further regularize $\mlp$ to map similar features to similar probabilities, we minimize its Lipschitz constant via $\loss_{reg}$ as detailed in~\cite[Eq.~(13)]{Liu2022}.
We then use $\mask^{\step}_\text{mlp}$, instead of $\mask^{\step}$, as inlier/outlier mask to train 3DGS in~\cref{eq:nvs}. 
We designate this configuration as `\methodsdmlp'. 
As we are \textit{co-training} our classifier together with the 3DGS model, additional care is needed in its implementation; see~\Cref{sec:sched}.

\subsection{Adapting 3DGS to robust optimization}
\label{sec:adaptations}
Directly applying any robust masking techniques to 3DGS can result in the robust mask overfitting to a premature 3DGS model (\cref{sec:sched}), \at{with inlier estimator becoming skewed by image-based training}~(\cref{sec:histogram}), or the densification tactics (\cref{sec:pruning}) of 3DGS. 
We propose solutions to these issues in what follows.

\subsubsection{Warm up with scheduled sampling}
\label{sec:sched}
We find it important to apply masks gradually, because the initial residuals are random.
This is doubly true if we use the learned clustering for masking since the MLP will not have converged early in the optimization, and predicts random masks.
Further, direct use of the outlier mask tends to  quickly overcommit to outliers, preventing valuable error back-propagation and learning from those regions.
We mitigate this by formulating our masking policy for each pixel as sampling from a Bernoulli distribution based on the masks:
\begin{equation}
    \mask^{\step} \sim \mathcal{B}\left( \sched \cdot 1 + (1-\sched) \cdot \mask^{\step}_{*}\right)\, ; 
\label{eq:schedule}
\end{equation}
where $\sched$ is a staircase exponential scheduler (detailed in the supplementary material~\ref{supp:scheduler}), going from one to zero, providing a warm-up.
This allows us to still sparsely sample gradients in areas we are not confident about, leading to better classification of outliers.

\subsubsection{Trimmed estimators in image-based training}
\label{sec:histogram}
As~\cite{Sabour2023robustnerf} implements a \textit{trimmed} estimator, the underlying assumption is that each minibatch (on average) contains the same proportion of outliers.
This assumption is broken in a 3DGS training run, where each minibatch is a \textit{whole} image, rather than a random set of  pixels drawn from the set of training images.
This creates a challenge in the implementation of the generalized median of~\cref{eq:robustnerf}, as the distribution of outliers is skewed between images.

We overcome this by tracking residual magnitudes over multiple training batches.
In particular, we discretize residual magnitudes into $B$ histogram buckets of width equal to the lower bound of rendering error ($10^{-3}$).
We update the likelihood of each bucket at each iteration via a discounted update to the bucket population, similar to fast median filtering approaches~\cite{medianfilter}.
This maintains a moving estimate of residual distribution, with constant memory consumption, from which we can extract our generalized median value~$\rho$ as the $\tau$ quantile in the histogram population; we refer the reader to our source code for implementation details.

\begin{figure*}[ht]
\includegraphics[width=\linewidth]{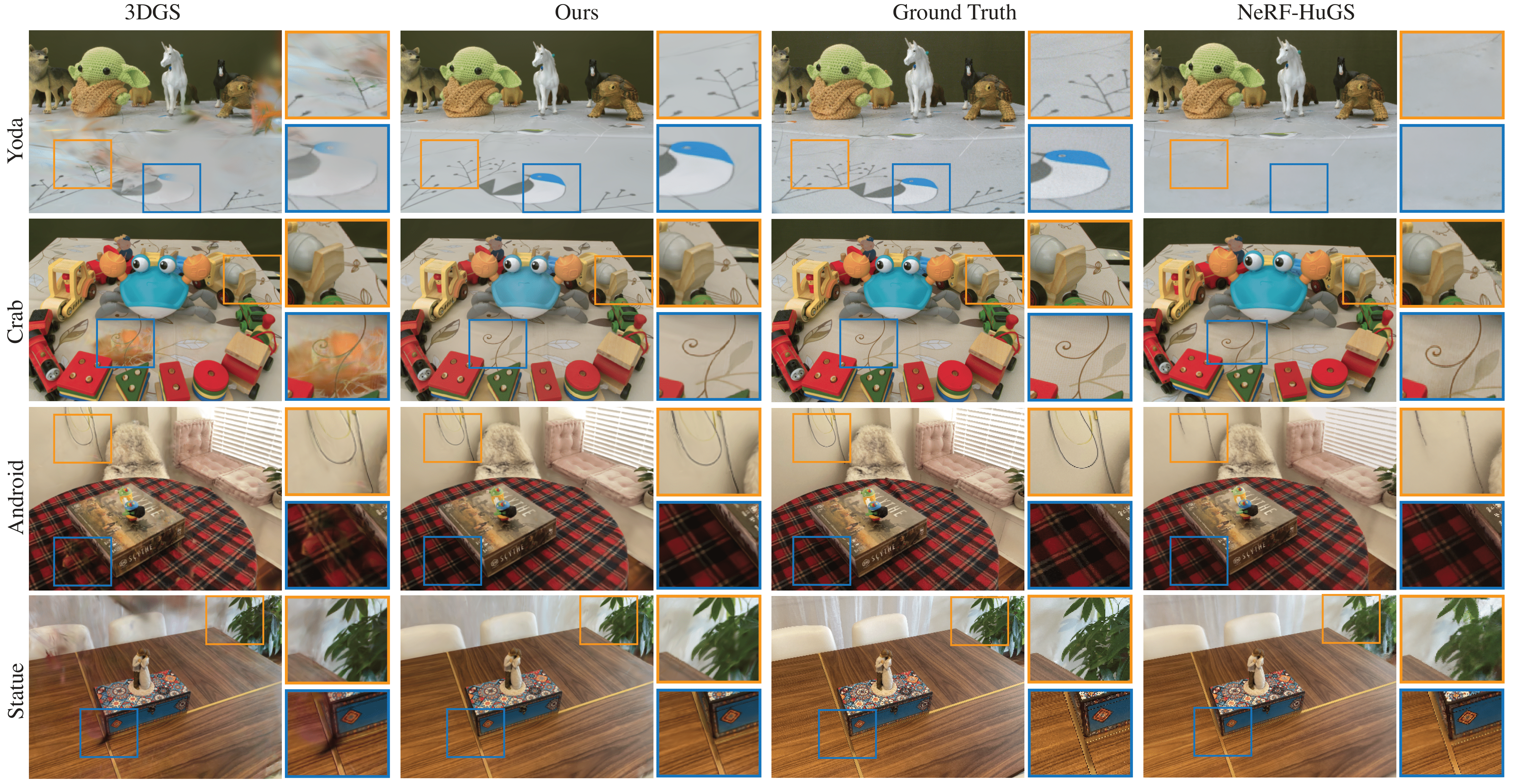}

\vspace*{0.2cm}
\setlength\tabcolsep{7.5pt}
\resizebox{\linewidth}{!}{
\begin{tabular}{l|ccc|ccc|ccc|ccc}
& \multicolumn{3}{c|}{Statue} & \multicolumn{3}{c|}{Android} & \multicolumn{3}{c|}{Yoda} & \multicolumn{3}{c}{Crab (1)} \\
 & \psnr & \ssim & \lpips & \psnr & \ssim & \lpips & \psnr & \ssim & \lpips & \psnr & \ssim & \lpips \\ \hline
MipNerf360 & 19.86 & .69 & .23 & 21.81 & .69 & .18 & 23.75 & .77 & .22 & 29.25 & .92 & .09  \\
RobustNerf & 20.60 & .76 & .15 & 23.28 & .75 & .13 & 29.78 & .82 & .15 & 32.22 & .94 & .06  \\
NeRF On-the-go & 21.58 & .77 & .24 & 23.50 & .75 & .21 & 29.96 & .83 & .24 & - & - & - \\
NeRF-HuGS & 21.00 & .77 & .18$^\dagger$ & 23.32 & .76 & .20$^\dagger$ & 30.70 & .83 & .22$^\dagger$ & 34.16 & .96 & \textbf{.07$^\dagger$} \\ \hline
\tdgsshortname & 21.68 & .83 & .14 & 23.33 & .80 & .15 & 27.15 & .92 & .13 & 31.80 & .96 & .08\\
{Our \methodsdmlp} & \textbf{22.69} & \textbf{.85} & \textbf{.12} & \textbf{25.15} & \textbf{.86} & \textbf{.09} & \textbf{33.60} & \textbf{.96} & \textbf{.10} & \textbf{35.85} & \textbf{.97} & .08 \\ \hline \hline
{\tdgsshortname} on \clean  & 28.02 & .95 & .05 & 25.42 & .87 & .07 & 33.69 & .94 & .12 & - & - & -\\
{\tdgsshortname}* on \clean & 28.63 & .95 & .04 & 25.38 & .87 & .07 & 36.34 & .97 & .07 & - & - & -
\end{tabular}
} %
\\[.75em]

\captionof{figure}{
Quantitative and qualitative evaluation on RobustNeRF~\cite{Sabour2023robustnerf} datasets show that SLS outperforms baseline methods on 3DGS and NeRF, by preventing over- or under-masking.
$\dagger$ denotes VGG LPIPS computed on NeRF-HuGS results rather than AlexNet LPIPS reported in NeRF-HuGS. 
{\tdgsshortname}* denotes \tdgsshortname with utility-based pruning.
}
\label{fig:robustnerf}
\end{figure*}

\subsubsection{A friendly alternative to ``opacity reset''}
\label{sec:pruning}
\cite{Kerbl2023tdgs} proposed to reset the opacity of all Gaussians every $M$ iterations. This opacity reset is a mechanism that deals with two main problems. 
First, in challenging datasets the optimization has the tendency to accumulate Gaussians close to the cameras. 
These are often referred to as \textit{floaters} in the literature.
Floaters are hard to deal with because they force camera rays to saturate their transmittance early and as a result gradients do not have a chance to flow through the occluded parts of the scene.
Opacity reset lowers the opacity of all Gaussians such that gradients can flow again along the whole ray. 
Second, opacity reset acts as a control mechanism for the number of Gaussians. 
Resetting opacity to a low value allows for Gaussians that never recover a higher opacity to be pruned by the adaptive density control mechanism~\cite{Kerbl2023tdgs}. 

However, opacity reset interferes with residual distribution tracking~(\cref{sec:histogram}), causing residuals to become artificially large in the iterations following opacity reset. 
Simply disabling does not work due to it's necessity to the optimization. 
Following~\cite{Goli2024bayesrays}, we instead propose utilization-based pruning (\methodprune).
We track the gradient of the rendered colors with respect to the projected splat positions\footnote{Please carefully note that this is the gradient of the rendered image with respect to Gaussian positions, and not the gradient of the loss.} $\x_g$ of each Gaussian $g$.
Computing the derivative with respect to projected positions, as opposed to 3D positions, allows for a less memory-intensive GPU implementation, while providing a similar metric as in~\cite{Goli2024bayesrays}. 
More concretely, we define the utilization as:
\begin{equation}
\util_{\element} =
\sum_{t \in \mathcal{N}_\steps(t)} 
\mathbb{E}_{w,h} \:
\left\|\mask^{(t)}_{h,w} \cdot \tfrac{ \partial \hat\image^{(t)}_{h,w}}{\partial\x^{(t)}_{\element}}\right\|_2^2
\label{eq:bayessplat}
\end{equation}
We average this metric across the image ($W{\times}H$), computing it every $\steps{=}100$ steps accumulated across the previous set of~$|\mathcal{N}_\steps(t)|{=}100$ images.
We prune Gaussians whenever $\util_{\element} {<} \pthresh$, with $\pthresh=10^{-8}$. 
Replacing opacity reset with utilization-based pruning achieves both original goals of opacity reset while alleviating interference to our residual distribution tracking. 
Utilization-based pruning significantly compresses scene representation by using fewer primitives while achieving comparable reconstruction quality even in outlier-free scenes; see \Cref{sec:exp_prune}. 
It also effectively deals with floaters; see \Cref{fig:ablation2}. 
Floaters, naturally, have low utilization as they participate in the rendering of very few views.
Furthermore, using masked derivatives as in~\cref{eq:bayessplat} allows for the removal of any splat that has leaked through the robust mask in the warm-up stage.

\subsubsection{Appearance modeling}
\label{sec:appearance}
While~\cite{Kerbl2023tdgs} assumed that the images of a scene~(up to distractors) are perfectly photometrically consistent, this is rarely the case for casual captures typically employing automatic exposure and white-balance.
We address this by incorporating the solution from~\cite{Rematas2022urf} adapted to the view-dependent colors represented as spherical harmonics from~\cite{Kerbl2023tdgs}.
We co-optimize a latent~$\aglolatent_{n} {\in} \mathbb{R}^{64}$ per input camera view, and map this latent vector via an MLP to a linear transformation acting on the harmonics coefficients~$\sphcoef$:
\begin{equation}
\hat{\sphcoef}_{i} = \mathbf{a} \odot \sphcoef_{i} + \mathbf{b},
\quad
\mathbf{a}, \mathbf{b} = \aglonet(\latent_n; \mlpparam_\aglonet)
\label{eq:glo}
\end{equation} 
where $\odot$ is the Hadamard product, $\mathbf{b}$ models changes in brightness, and $\mathbf{a}$ provides the expressive power for white-balance. 
During optimization, the trainable parameters also include $\mlpparam_\aglonet$ and $\{\latent_n\}$.
Such a reduced model can prevent $\latent_n$ from explaining distractors as per-image adjustments, as would happen in a simpler GLO~\cite{MartinBrualla21cvpr_nerfw}; see~\cite{Rematas2022urf} for an analysis.

\section{Results}
\label{sec:experiments}

In what follows, we compare our proposed method on established datasets of casual distractor-filled captures~(\cref{sec:exp_reconstruct}), comparing with other methods.
We then investigate the effect of our proposed opacity reset alternative pruning~(\cref{sec:exp_prune}). Finally, we report a complete analysis of different variants of our clustering, along with an ablation study of our design choices ~(\cref{sec:exp_ablation}).

\paragraph{Datasets}
We evaluate our method on the RobustNeRF~\cite{Sabour2023robustnerf} and NeRF on-the-go~\cite{Ren2024nerfonthego} datasets of \textit{casual captures}. 
The RobustNeRF dataset includes four scenes with distractor-filled and distractor-free training splits, allowing us to compare a robust model with a `clean' model trained on distractor-free images.
All models are evaluated on a clean test set.
The `Crab' and `Yoda' scenes feature variable distractors across images, not captured in a single casual video, but these exact robotic capture with twin distractor-free and distractor-filled images allow a fair comparison to the `clean' model.
Note the (originally released) Crab~(1) scene had a test set with same set of views as those in the train set, which is fixed in Crab~(2).
We compare previous methods on Crab~(1), and present full results on Crab~(2) in~\Cref{sec:exp_ablation}, and in the supplementary material~\ref{supp:crab}.
The NeRF on-the-go dataset has six scenes with three levels of transient distractor occlusion~(low, medium, high) and a separate clean test set for quantitative comparison.

\begin{figure}[t]
\centering
\setlength\tabcolsep{1.8pt}
\includegraphics[width=\linewidth]{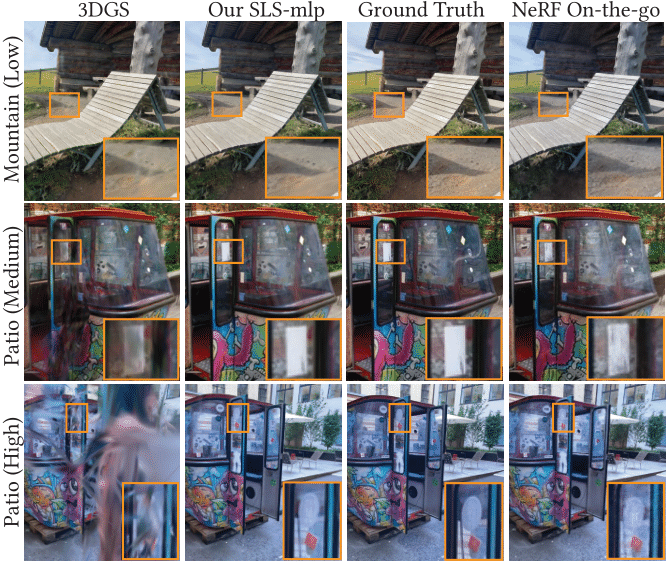}
\resizebox{\linewidth}{!}{

\begin{tabular}{l|ccc|ccc|ccc}
 & \multicolumn{3}{c|}{Mountain} & \multicolumn{3}{c|}{Patio} & \multicolumn{3}{c}{Patio High}\\
 & \psnr & \ssim & \lpips & \psnr & \ssim & \lpips & \psnr & \ssim & \lpips  \\ \hline
MipNerf360 & 19.64 & 0.60 & 0.35 & 15.48 & 0.50 & 0.42 & 15.73 & 0.43 & 0.49 \\
RobustNerf & 18.07 & 0.49 & 0.49 & 17.55 & 0.53 & 0.45 & 12.99 & 0.35 & 0.61 \\
Nerf On-the-go & 20.15 & 0.64 & 0.26 & 20.78 & 0.75 & 0.22 & 21.41 & 0.72 & 0.24 \\ \hline
\tdgsshortname & 20.18 & 0.70 & 0.23 & 18.25 & 0.71 & 0.23 & 18.14 & 0.68 & 0.30 \\
Our \methodsdmlp & \textbf{22.53} & \textbf{0.77} & \textbf{0.18} & \textbf{22.24} & \textbf{0.86} & \textbf{0.10} & \textbf{22.84} & \textbf{0.83} & \textbf{0.16}
\end{tabular}
} %
\vspace*{-.6em}
\captionof{figure}{
SLS reconstructs scenes from NeRF On-the-go~\cite{Ren2024nerfonthego} dataset in great detail. High-occlusion lingering distractors, lead to distractor leaks modeled as noisy floaters in baselines. Our method is free of such artifacts.
}
\vspace*{-0.2cm}
\label{fig:nerfonthego2}
\end{figure}
\paragraph{Baselines} 
Distractor-free reconstruction has yet to be widely addressed by 3D Gaussian Splatting methods.
Existing methods mostly focus on global appearance changes such as brightness variation~\cite{Dahmani2024swag, Wang2024wegs, Kulhanek2024wildgaussians}, and do not focus on the distractor-filled datasets of casual captures curated for this task.  
We therefore compare against vanilla 3DGS and robust NeRF methods. We further add GLO to the vanilla 3DGS baseline to be comparable with MipNeRF360 results that have GLO enabled.
We compare against state-of-the-art NeRF methods, NeRF on-the-go~\cite{Ren2024nerfonthego}, NeRF-HuGS~\cite{chen2024nerfhugs} and RobustNeRF~\cite{Sabour2023robustnerf}.
We also include MipNeRF-360~\cite{Barron2022mipnerf360} as a baseline for NeRF.

\paragraph{Metrics}
We compute the commonly used image reconstruction metrics of PSNR, SSIM and LPIPS.
We use normalized VGG features, as most do, when computing LPIPS metrics.
NeRF-HuGS~\cite{chen2024nerfhugs} reports LPIPS metrics from AlexNet features; for fair comparison, we compute and report VGG LPIPS metrics on their released renderings.
Finally, note NeRF on-the-go does not evaluate on `Crab', because of the aforementioned issue.

\begin{figure}[t]
\includegraphics[width=\linewidth]{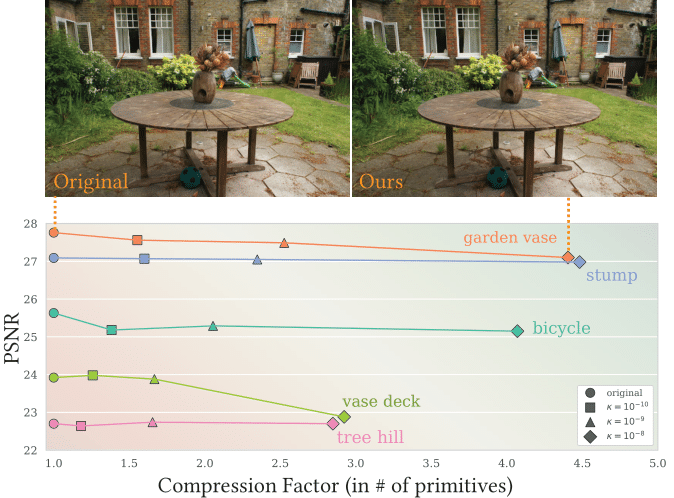}
\vspace*{-0.5cm}
\captionof{figure}{
Quantitative and qualitative results on MipNeRF360~\cite{Barron2022mipnerf360} dataset shows gradient-based pruning can reduce the number of Gaussians up to $4.5\times$ with only marginal degradation of image quality. 
}
\vspace*{-0.3cm}
\label{fig:pruning}
\end{figure}

\begin{figure*}[ht]
\centering
\setlength\tabcolsep{1.8pt}
\includegraphics[width=\linewidth]{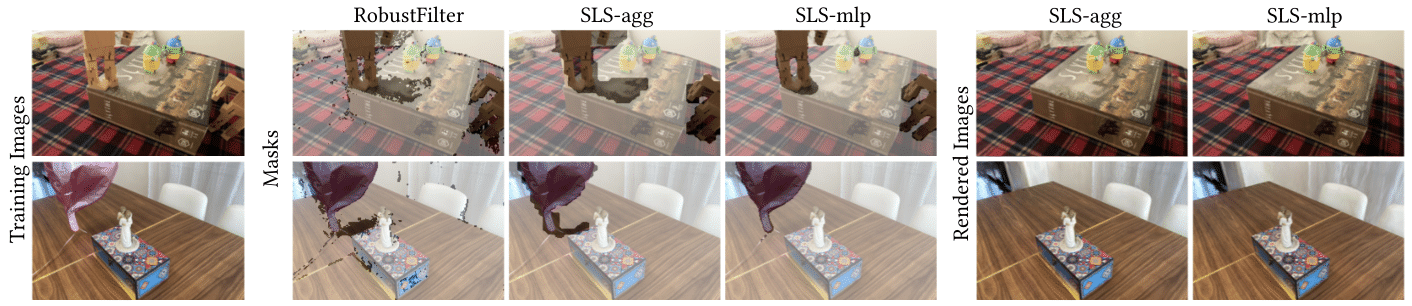}

\resizebox{\linewidth}{!}{
\begin{tabular}{l|cccc|cccccc|c}
 & Android & Statue & Crab (2) & Yoda & Mountain & Fountain & Corner & Patio & Spot & Patio-High & Average \\ \hline
\tdgsshortname  & $23.33 \pm 0.13$ & $21.68 \pm 0.16$ & $29.74 \pm 0.37$ & $27.15 \pm 0.61$ & $20.90 \pm 0.18$ & $21.85 \pm 0.27$ & $23.39 \pm 0.43$ & $18.33 \pm 0.27$ & $21.50 \pm 0.85$ & $18.06 \pm 0.71$ & $22.60$ \\
\methodfilter  & $24.50 \pm 0.05$ & $22.70 \pm 0.06$ & $31.34 \pm 0.13$ & $33.23 \pm 0.13$ & $22.29 \pm 0.07$ & $22.59 \pm 0.07$ & $25.20 \pm 0.10$ & $18.16 \pm 0.19$ & $25.54 \pm 0.08$ & $23.01 \pm 0.18$ & $24.85$ \\
\methodsdcluster & $24.94 \pm 0.08$ & $\mathbf{23.16 \pm 0.08}$ & $33.50 \pm 0.14$ & $35.01 \pm 0.21$ & $22.65 \pm 0.14$ & $23.03 \pm 0.17$ & $26.33 \pm 0.10$ & $22.31 \pm 0.13$ & $\mathbf{26.34 \pm 0.37}$ & $\mathbf{23.54 \pm 0.15}$ & $26.08$ \\
\methodsdmlp w/o UBP & $25.08 \pm 0.04$ & $22.75 \pm 0.14$ & $\mathbf{34.43 \pm 0.03}$ & $\mathbf{34.36 \pm 0.24}$ & $\mathbf{22.93 \pm 0.09}$ & $\mathbf{23.19 \pm 0.13}$ & $\mathbf{26.74 \pm 0.13}$ & $\mathbf{22.36 \pm 0.07}$ & $25.95 \pm 0.47$ & $23.27 \pm 0.13$ & $\mathbf{26.11}$ \\
\hline
\methodsdmlp w/ \methodprune& $\mathbf{25.15 \pm 0.05}$ & $22.69 \pm 0.16$ & $33.63 \pm 0.27$ & $33.60 \pm 0.30$ & $22.53 \pm 0.11$ & $22.81 \pm 0.10$ & $26.43 \pm 0.08$ & $22.24 \pm 0.19$ & $25.76 \pm 0.15$ & $22.84 \pm 0.32$ & $25.77$

\end{tabular}
}
\captionof{figure}{
We ablate our different robust masking methods on~\cite{Sabour2023robustnerf} and~\cite{Ren2024nerfonthego} datasets. The reconstruction metrics and qualitative masks illustrate the performance of \methodsdcluster~\cref{eq:agglomerative} and \methodsdmlp~\cref{eq:learnt} over a basic \methodfilter~\cref{eq:robustnerf} adapted from~\cite{Sabour2023robustnerf}, and baseline vanilla 3DGS\cite{Kerbl2023tdgs}.
The final row enables Utility-Based Pruning (\methodprune) (\cref{sec:pruning}).
All methods use opacity reset disabled, the same scheduling in \cref{eq:schedule}, and GLO \cref{eq:glo} enabled on all runs including \tdgsshortname.
\methodsdcluster and \methodsdmlp are mostly within $2\sigma$ of each other on all tasks.
The $\sigma$ is calculated from $5$ independent runs each. 
}
\label{fig:robustprogress}
\end{figure*}

\paragraph{Implementation details} 
We train our 3DGS models with the same hyper-parameters as in the officially released codebase.
All models are trained for 30k iterations. 
We turn off the opacity-reset and only reset the non-diffuse spherical harmonic coefficients to 0.001 at the 8000th step.
This ensures that any distractors leaked in the earlier stages of MLP training do not get modeled as view dependent effects.
We run \methodprune every 100 steps, from the 500th to 15000th step.
For MLP training, we use the Adam optimizer with a 0.001 learning rate. 
We compute image features from the 2nd upsampling layer of Stable diffusion v2.1, denoising time step of 261, and an empty prompt.
\cite{Tang2023emergent} found this configuration most efficient for segmentation and keypoint correspondence tasks.
We concatenate positional encoding of degree 20 to the features as input to the MLP. 

\subsection{Distractor-free 3D reconstruction}
\label{sec:exp_reconstruct}

We evaluate our method by preforming 3D reconstruction on the RobustNeRF and NeRF on-the-go datasets. 
In \Cref{fig:robustnerf}, we quantitatively show that \methodsdmlp outperforms all the robust NeRF-based baselines on the RobustNeRF dataset.
The results further show that we improve significantly upon vanilla 3DGS, while having closer performance to the ideal clean models, specifically on `Yoda' and `Android'. We further qualitatively compare with vanilla 3DGS and NeRF-HuGS.
The qualitative results show that vanilla 3DGS tries to model distractors as noisy floater splats (`Yoda', `Statue') or view-dependent effects (`Android') or a mixture of both (`Crab'). 
NeRF-HuGS~\cite{chen2024nerfhugs} which uses segmentation-based masks shows signs of over-masking (removing static parts in all four scenes), or under-mask in challenging sparsely sampled views letting in transient objects (`Crab'). 

In \Cref{fig:nerfonthego1} and \Cref{fig:nerfonthego2}, we perform a similar analysis on the NeRF On-the-go~\cite{Ren2024nerfonthego} dataset. 
While we show superior quantitative results to both SOTA robust NeRF methods, we also achieve a significant performance boost compared to vanilla 3DGS. The results further show that for low occlusion scenes the robust initialization of vanilla 3DGS from COLMAP~\cite{Schonberger2016sfmrevisited} point clouds, specifically RANSAC's rejection of outliers, is enough to yield good reconstruction quality.
However, as the distractor density increases, 3DGS reconstruction quality drops with qualitative results showing leakage of distractor transients.
Additionally, qualitative results show that NeRF On-the-go fails to remove some of the distractors included in the early stages of training (`Patio', `Corner', `mountain' and `Spot'), showing further signs of overfitting to the rendering error.
This also is seen in the over-masking of fine details (`Patio High') or even bigger structures (`Fountain') removed completely.

\subsection{Effect of utilization-based pruning}
\label{sec:exp_prune}
\begin{figure}[t]
\centering
\setlength\tabcolsep{1.8pt}
\includegraphics[width=\linewidth]{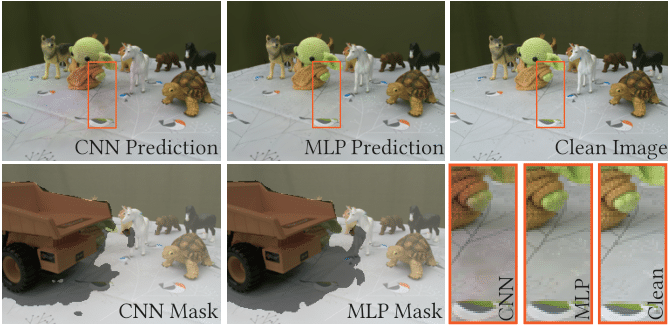}\\
\vspace*{0.1cm}
\resizebox{\linewidth}{!}{
\begin{tabular}{c|c|c|c|c||c|c|c}
                   & \begin{tabular}[c]{@{}c@{}}SLS- \\mlp\end{tabular} & \begin{tabular}[c]{@{}c@{}}CNN \\\end{tabular}    & \begin{tabular}[c]{@{}c@{}}SLS-mlp \\ ($\lambda=0.1)$\end{tabular} & \begin{tabular}[c]{@{}c@{}}SLS-mlp \\ ($\lambda=1)$\end{tabular} & \begin{tabular}[c]{@{}c@{}} SLS- \\ agg\end{tabular}   & \begin{tabular}[c]{@{}c@{}}SLS-agg \\ ($\numclusters=10$)\end{tabular} & \begin{tabular}[c]{@{}c@{}}SLS-agg \\ ($\numclusters=1000$)\end{tabular} \\ \hline
PSNR $\uparrow$    & 28.72 & 27.91 & 28.71                                                        & 28.61                                                      & 28.91 & 27.71                                                 & 29.00                                                   \\ \hline
SSIM $\uparrow$    & 0.90  & 0.89  & 0.90                                                         & 0.89                                                       & 0.90  & 0.89                                                  & 0.90                                                    \\ \hline
LPIPS $\downarrow$ & 0.10  & 0.13  & 0.11                                                         & 0.12                                                       & 0.10  & 0.11                                                  & 0.09                                                   
\end{tabular}
}%
\vspace*{-.5em}
\captionof{figure}{
Ablations on variants from \Cref{sec:semantic} show
replacing the MLP \cref{eq:learnt} in \methodsdmlp with a CNN reduces quality. 
Varying its regularization coefficient $\lambda$ in \cref{eq:mlploss} shows minimal impact. 
More agglomerative clusters in \methodsdcluster \cref{eq:agglomerativeprob} improve performance, plateauing for $\numclusters{\geq}100$. Metrics averaged over all RobustNeRF dataset.
}
\label{fig:ablation1}
\end{figure}
\begin{figure}[t]
\centering
\setlength\tabcolsep{5.8pt}
\includegraphics[width=\linewidth]{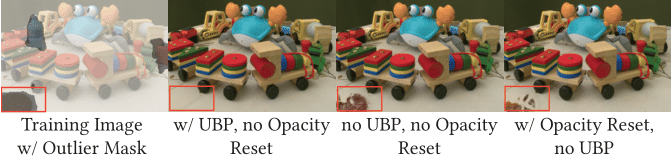}\\
\vspace*{0.1cm}
\resizebox{\linewidth}{!}{
\begin{tabular}{c|c|c|c|c|c|c}
                   & SLS-mlp   & No \methodprune & Opacity Reset & No GLO & No $\alpha$ & No $\mathcal{B}$ \\ \hline
PSNR $\uparrow$    & 28.72 & 29.27    & 27.64         & 28.43  & 28.41       & 28.41            \\ \hline
SSIM $\uparrow$    & 0.90  & 0.90     & 0.89          & 0.89   & 0.90        & 0.90             \\ \hline
LPIPS $\downarrow$ & 0.10  & 0.09     & 0.11          & 0.11   & 0.11        & 0.11            
\end{tabular}
}%
\vspace*{-.5em}
\captionof{figure}{
Ablation on adaptations from \Cref{sec:adaptations} show
disabling \methodprune (\cref{sec:pruning}) may produce higher reconstruction metrics but leaks transients as seen in the lower-left corner of the image; replacing it with ``Opacity Reset'' as originally introduced in 3DGS is also ineffective.
GLO appearance modelling \cref{eq:glo} improves quality, as do scheduling~($\alpha$), and Bernoulli sampling ($\mathcal{B}$)~\cref{eq:schedule}. Experiments are executed on \methodsdmlp, with metrics averaged over all RobustNeRF dataset.
}
\vspace*{-0.55cm}
\label{fig:ablation2}
\end{figure}
In all our experiments, enabling our proposed utilization-based pruning (\methodprune) (\cref{sec:pruning}), decreases the number of Gaussians from $4\times$ to $6\times$.
This compression translates to at least a $2\times$ reduction in training time with \methodprune enabled and $3\times$ during inference.
\Cref{fig:ablation2} shows that enabling \methodprune may degrade quantitative measurements slightly, but in practice the final renderings are cleaner with less floaters (e.g. bottom left of the image).
Similar observations indicate that metrics such as PSNR and LPIPS may not completely reflect the presence of floaters as clearly as a rendered video.
Given the substantial reduction in number of Gaussians, we propose  \methodprune as a compression technique applicable to cluttered, \emph{as well as clean}, datasets. \Cref{fig:pruning} shows that on clean MipNeRF360 \cite{Barron21iccv_Mip_NeRF} datasets, using \methodprune instead of opacity reset reduces the number of Gaussians from $2\times$ to $4.5\times$ while preserving rendering quality.

\subsection{Ablation study}
\label{sec:exp_ablation}
In \Cref{fig:robustprogress}, we compare the performance of \methodshortname with a progression of other robust masking techniques.
The progression begins with a naive application of a robust filter~\eqref{eq:robustnerf}, followed by the application of \methodsdcluster, and finally the use of an MLP in \methodsdmlp.
We demonstrate that both \methodsdcluster and \methodsdmlp are capable of effectively removing distractors from the reconstructed scene, while maintaining maximal coverage of the scene.
Further, in~\Cref{fig:ablation1} and~\Cref{fig:ablation2} we ablate our choices in both architectural design, and the adaptations proposed in~\Cref{sec:adaptations}.
\cref{fig:ablation1} shows that using an MLP instead of a small CNN (both roughly having 30K parameters, and two non-linear activations) can adapt better to subtle transients like shadows.
The choice of regularizer weight $\lambda$ seems to have little effect. 
In agglomerative clustering, more clusters generally lead to better results, with gains diminishing after 100 clusters.
\Cref{fig:ablation2} further illustrates the effectiveness of \methodprune in removing leaked distractors.
Our other adaptations, GLO, warm-up stage and Bernoulli sampling all show improvements.

\section{Conclusion}
We have presented \methodname, a method for transient distractor suppression for 3DGS.
We established a class of masking strategies that exploit semantic features to effectively identify transient distractors without any explicit supervision. 
Specifically, we proposed a spatial clustering method `\methodsdcluster' that is fast and does not require further training, simply assigning an inlier-outlier classification to each cluster.
We then proposed a spatio-temporal learned clustering based on training a light-weight MLP simultaneously with the 3DGS model, `\methodsdmlp', that allows for higher precision grouping of semantically associated pixels, while marginally slower than clustering.
Our methods leverage the semantic bias of Stable Diffusion features and robust techniques to achieve state of the art suppression of transient distractors.
We also introduced a gradient-based pruning method that offers same reconstruction quality as vanilla 3DGS, while using significantly lower number of splats, and is compatible with our distractor suppression methods.
We believe that our work is an important contribution necessary for widespread adoption of 3DGS to real-world in-the-wild applications.

\paragraph{Limitations} 
Our reliance on text-to-image features, although generally beneficial for robust detection of distractors, imposes some limitations.
One limitation is that when distractor and non-distractors of the same semantic class are present and in close proximity, they may not be distinguished by our model. Details are discussed further in the supplementary materials~\ref{supp:similarity}.
Further, the low-resolution features these models provide can miss thin structures such as the balloon string of \Cref{fig:robustprogress}.
Especially in the use of clustering, upsampling the features to image resolution results in imprecise edges. 
Our pruning strategy, is based on epistemic uncertainty computation per primitive which is effective in removing lesser utilized Gaussians. However if the uncertainty is thresholded too aggressively (e.g. `vase deck' in~\cref{fig:pruning}), it can remove parts of the scene that are rarely viewed in the training data. 

\begin{acks}
We thank Abhijit Kundu, Kevin Murphy, Songyou Peng, Rob Fergus and Sam Clearwater for reviewing our manuscript and for their valuable feedback.
\end{acks}

\bibliographystyle{ACM-Reference-Format}
\bibliography{nerf, non-nerf}

\clearpage
\appendix
\section*{Supplementary Material}

\section{Distinguishing between semantically similar instances}
\label{supp:similarity}
\begin{figure}[t]
    \centering
    \includegraphics[width=\linewidth]{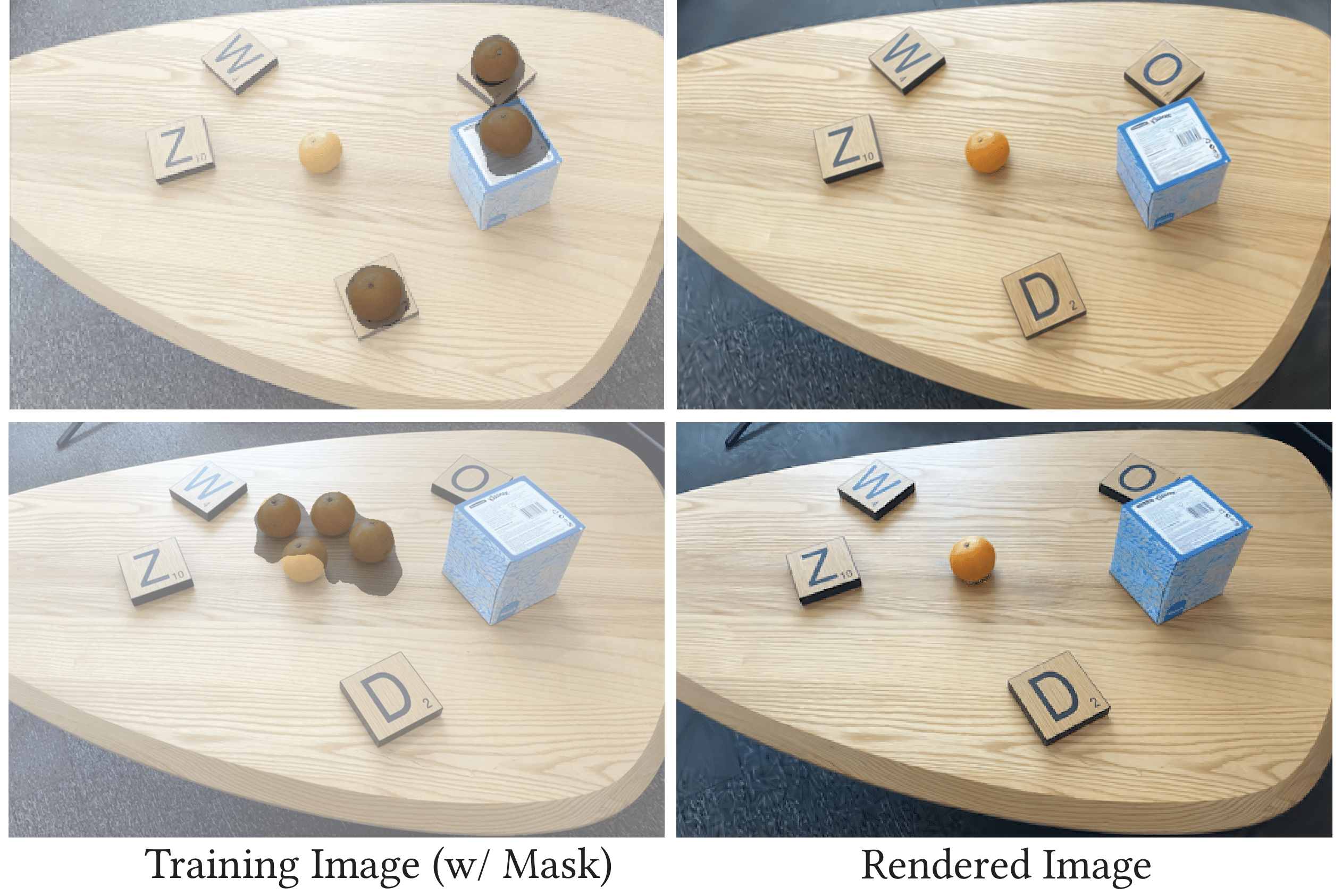}
    \caption{SLS-MLP can correctly distinguish between similar-looking oranges when the non-dsitractor instances are far from the distractor instance (centered on the table). However, the masking task becomes more difficult for nearby instances of distractor and non-distractors.}
    \label{fig:oranges}
\end{figure}

Our method relies on features extracted from text-to-image diffusion models, to reliably learn the subspace of features that represent distractors appearing in a casual capture. However, this implies that in cases where similar instances of a semantic class appear as both distractors and non-distractors, our model would not be able to distinguish the two. We show in~\cref{fig:oranges} that this is not generally true. When the instances of the same class are not very close in image space, the model can distinguish between distractor and non-distractor orange instances, however in scenarios where they are very close we can see over-masking of the non-distractor orange. We hypothesize that this is a result of foundation features encoding not only semantics, but also appearance and position in the image as shown in~\cite{El_Banani_2024_CVPR}.

\section{Warm-up scheduler}
\label{supp:scheduler}
As explained in~\cref{sec:adaptations}, we use a staircase exponential scheduler for our warm up phase:
\begin{equation}
    \sched = \exp \left(-\beta_1 \left\lfloor \frac{(t + 1)}{\beta_2} \right\rfloor \right),
\end{equation}
where $t$ is the time step in optimization, $\beta_1$ controls the speed of decay and $\beta_2$ determines the length of steps in the staircase function. We use $\beta_1 = 3\times10^{-4}$ and $\beta_2 = 1.5$ for all experiments, but the three highest occlusion rate scenes in NeRF On-the-go~\cite{Ren2024nerfonthego} dataset where we use $\beta_1 = 3\times10^{-3}$ for a fastest decay.

\section{Additional results on the Crab dataset}
\label{supp:crab}
The Crab dataset in RobustNeRF~\cite{Sabour2023robustnerf} has two released versions, one without any additional viewpoints for testing and one with an extra test set of camera viewpoints. In the main paper we refer to the first version as Crab (1) and the second version as Crab (2). While previous work has only tested on Crab (1), Crab (2) has the conventional format of NeRF datasets with separate test views. We present additional results on Crab (2) scene in~\cref{tab:crab2}, showing that SLS-MLP has a very close performance to the ideal 3DGS model trained on clean data, when tested from different viewpoints than training datset.
\begin{table}[h]
\centering
\begin{tabular}{c|c|c||c|c}
      & SLS-MLP & 3DGS  & 3DGS Clean & 3DGS* Clean \\ \hline
PSNR $\uparrow$ & \textbf{34.35}   & 26.33 & 33.43         & \textbf{35.58}          \\ \hline
SSIM $\uparrow$ & \textbf{0.96}    & 0.91  & 0.94          & \textbf{0.97}           \\ \hline
LPIPS $\downarrow$ & \textbf{0.03}    & 0.08  & 0.05          & \textbf{0.01}          
\end{tabular}
\captionof{table}{Quantitative result on Crab (2) dataset, where a test set with different viewpoints than training is provided, shows superior performance of SLS-MLP to vanilla 3DGS and close performance to the ideal model trained on clean data. 3DGS* denotes use of utility-based pruning.}
\label{tab:crab2}
\end{table}

\section{Additional Results on NeRF On-the-go dataset}
\begin{figure*}
    \centering
    \includegraphics[width=\linewidth]{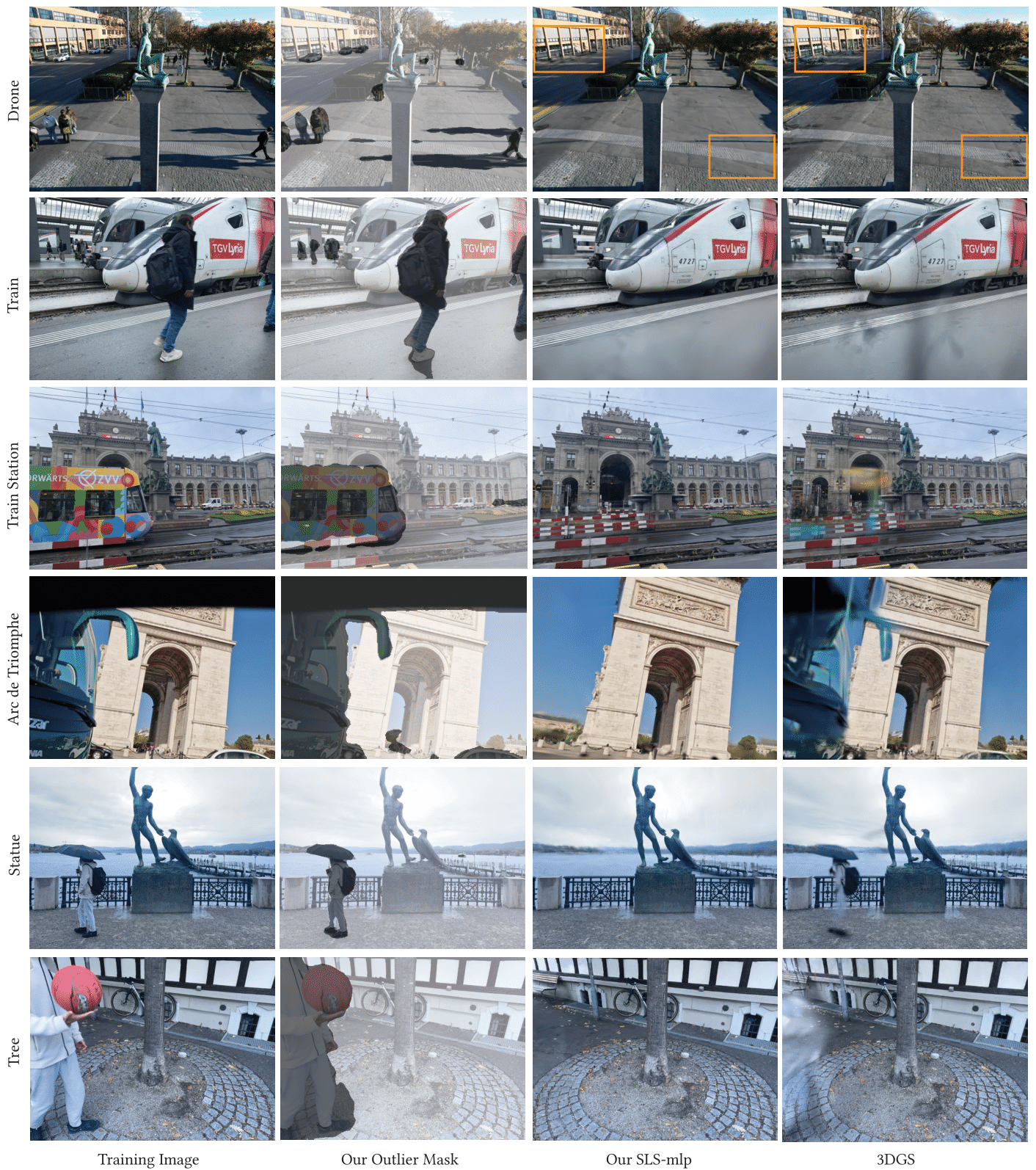}
    \captionof{figure}{Qualitative results on scenes from NeRF On-the-go~\cite{Ren2024nerfonthego} show robustness of our method to transparent surfaces such as glass windshields (Train Station, Arc de Triomphe), and similarly-colored distractors and backgrounds (Tree, Statue). Further, our method shows robustness to distractor shadows to a degree (Drone, Train).}
    \label{fig:supp}
\end{figure*}

NeRF On-the-go dataset~\cite{Ren2024nerfonthego} provides six additional scenes for qualitative evaluation only. We provide qualitative results on these scenes in~\cref{fig:supp}. In the `Drone' scene, our method is able to detect harder shadows of people and remove them seamlessly. However, complete robustness to softer shadows is a limitation of our work, as the semantic class of shadows is not reflected very well in the text-to-image features that we use. This can be seen in the `Train' scene, where shadows of people are only detected to a degree. Further, in the`Train Station' and `Arc de Triomphe' scenes we see that our model shows robustness to transparent surfaces on distractors such as glass windshields. Finally, in the `Statue' and `Tree' scenes, SLS-MLP works well in distinguishing between the distractor and the background, even though the distractors (mostly) have very similar color to their background.

\end{document}